# Dual-view Spatio-Temporal Feature Fusion with CNN-Transformer Hybrid Network for Chinese Isolated Sign Language Recognition

Siyuan Jing[1,2†*], Guangxue Wang[3†], Haoyang Zhai[3], Qin Tao[1,2], Jun Yang[1,2], Bing Wang[3], Peng Jin[1,2*]

(1) Sichuan Province Key Laboratory of Philosophy and Social Science for Language Intelligence in Special Education, Leshan Normal University, Leshan China, 614000

(2) Key Laboratory of Internet Natural Language Intelligent Processing of Sichuan Provincial Education Department, Leshan Normal University, Leshan China, 614000

(3) School of Computer Sciences and Software Engineering, Southwest Petroleum University, Chengdu China, 610000

†Co-first author
*Co-corresponding author

**Abstract:** Due to the emergence of many sign language datasets, isolated sign language recognition (ISLR) has made significant progress in recent years. In addition, the development of various advanced deep neural networks, e.g., CNNs, GCNs, and transformers, is another reason for this breakthrough. However, challenges remain in applying the technique in the real world. First, existing sign language datasets do not cover the whole sign vocabulary, e.g., there are 8124 words in Chinese national sign vocabulary, but the largest CSL (Chinese Sign Language) dataset, i.e., DEVISIGN, contains only 2000 words. Second, most of the sign language datasets provide only single-view RGB videos, which makes it difficult to handle hand occlusions when performing ISLR. To fill this gap, this paper presents a dual-view sign language dataset for ISLR named NationalCSL-DP, which fully covers the Chinese national sign language vocabulary. The dataset consists of 134k+ sign videos recorded by ten signers with respect to two vertical views, namely, the front side and the left side. Furthermore, a CNN–transformer network is also proposed as a strong baseline and an extremely simple but effective fusion strategy for prediction. Extensive experiments were conducted to prove the effectiveness of the datasets as well as the baseline. The results show that the proposed fusion strategy can significantly increase the performance of the ISLR, but it is not easy for the sequence-to-sequence model, regardless of whether the early-fusion or late-fusion strategy is applied, to learn the complementary features from the sign videos of two vertical views.

**Keywords:** Isolated sign language recognition; dataset; deep neural network; dual-view fusion; Chinese national sign language vocabulary

## I. Introduction

Sign language is the primary tool of communication for approximately 400 million individuals who are deaf or hard of hearing. Sign language is a vision language that relies on hand gestures, facial expressions, and body movements to convey complex thoughts and emotions [1]. Learning sign language is a rather difficult task for people without hearing impairments since sign language from different regions may differ significantly, even between two close cities. To address this issue, sign language recognition (SLR) has attracted considerable attention from the academic community over the last decade. SLR can be divided into isolated sign language recognition (ISLR) [2] and continuous sign language recognition (CSLR) [3] (also known as sign language translation [4-6]) according to whether the object being recognized is a word or a sentence. In this paper, we focus on ISLR.

ISLR aims to decode a word-level sign signal into a gloss, which serves as a crucial tool for bridging the communication gap between hearing and nonhearing communities. Most ISLR approaches rely on (RGB or depth) cameras or wearable devices (such as sensors or colored gloves). However, the latter is intrusive, which makes deaf individuals uncomfortable and thus limits its applicability. For this reason, researchers have recently paid more attention to the purely vision-based ISLR solution. However, extracting the fine-grained action semantic features in sign videos is a challenging task.

With the rapid development of deep neural networks and the emergence of numerous sign language datasets, video-based ISLR has made significant progress in the past decade. At the technical level, various deep neural networks (DNNs), such as convolutional neural networks (CNNs), graph convolutional networks (GCNs) and transformers, increase the performance of ISLR, as they can extract spatial–temporal features effectively from sign videos. At the data level, researchers have constructed large-scale sign language datasets to support the training of DNN models for ISLR. However, challenges remain in applying the ISLR technique in the real world. First, existing sign language datasets do not cover the whole sign vocabulary, e.g., there are 8124 words in the Chinese national sign vocabulary (CNSV), but the largest Chinese Sign Language (CSL) dataset, i.e., DEVISIGN [7], contains only 2000 words. Second, most of the sign language datasets provide only single-view RGB videos, which makes it difficult to handle hand occlusions when performing ISLR. An example is given in Figure 1, in which a signer is signing the word "air-conditioner". The front side and the left side of the signer are both shown. The signer's hands are occluded in the front-side video; therefore, extracting the features of the hands, including the shape and movement, is difficult. However, from the left side, we can clearly distinguish the signer's two hands.

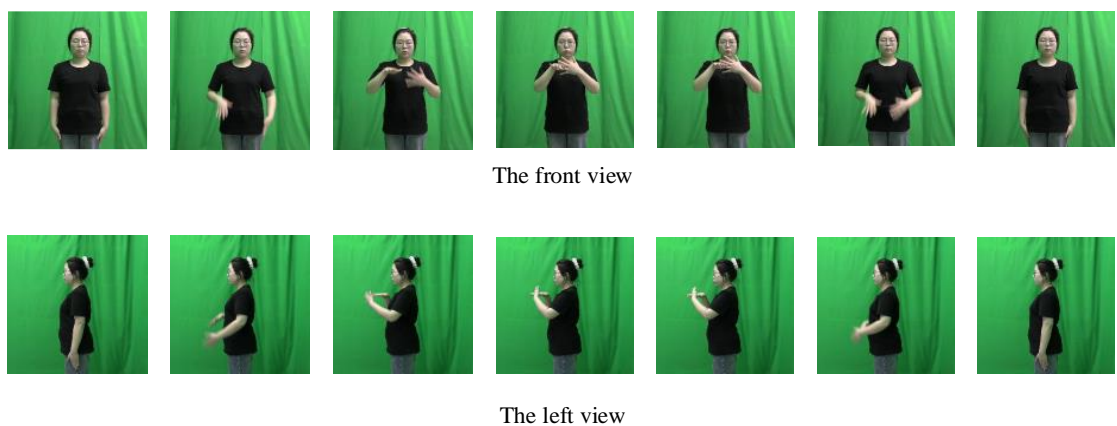

The front view

The left view

Figure1. Signs of the word "air-conditioner" from two perspectives

To address the above issues, this paper presents a new Chinese ISLR dataset named NationalCSL-DP (DP means dual perspectives), which contains the most extensive vocabulary compared with the existing public ISLR datasets. Specifically, we asked CSL experts to exclude glosses with the same sign from the CNSV. For example, "战斗" (combat), "战役" (campaign) and "战争" (war) are three different glosses but with the same sign, and we keep the gloss "战斗", which first appears in the CNSV. Finally, we obtained 6707 glosses. After approximately five months of preparation and recording, we captured more than 131K+ high-quality sign videos with the assistance of 10 signers. Each video recording was supervised by at least one CSL expert to ensure the precision of the sign language expression. The videos were recorded at two green screen studios. Both studios were equipped with two high-definition cameras, which were placed on the front and left sides of the signer. All the cameras recorded videos at 1920 × 1080 resolution and 50 fps. To our knowledge, the NationalCSL-DP dataset is the first dataset for isolated Chinese Sign Language recognition that provides aligned dual-view videos of sign language and covers all the signs in the CNSV. In this way, it enables more comprehensive training and testing of ISLR models. In addition, we propose a strong baseline on NationalCSL-DP that employs a CNN−transformer as the backbone network. The CNN module extracts spatial features from sign videos, and the transformer module models the temporal features. A simple but efficient fusion strategy is adopted.

The contributions of this work are as follows:

(1) A new ISLR dataset named NationalCSL-DP, which consists of 131K+ high-quality sign videos and covers 6707 glosses, is released. To our knowledge, this is the first ISLR dataset that covers all the signs in the CNSV, and it provides aligned dual-view sign videos that aim to handle the hand occlusions.

(2) A strong baseline employing CNN−transformer as the backbone network is also proposed. Moreover, we tested three popular fusion strategies, namely, early fusion, late fusion and plus fusion, to predict the glosses of sign videos.

(3) Following some popular ISLR datasets, the NationalCSL-DP dataset is arranged into five datasets with different scales, including NationalCSL200, NationalCSL500, NationalCSL1000, NationalCSL2000 and NationalCSL6707. Extensive experiments were conducted on these datasets to evaluate the quality of the dataset as well as the effectiveness of the baseline.

The rest of the paper is organized as follows. Section 2 reviews the related work. Section 3 explains the details of the NationalCSL-DP dataset. Section 4 introduces a baseline for the dataset, which is based on the CNN−transformer architecture. Section 5 presents the experimental results and provides an in-depth analysis. Finally, Section 6 concludes the work.

## II. Related works

### 2.1 Isolated sign language recognition

This section focuses only on ISLR technologies based on computer vision. We can classify ISLR technologies into two categories, namely, nondeep learning methods and deep learning methods. The former usually uses computer vision methods to extract features of sign language from videos, such as the scale-invariant feature transform (SIFT), histogram of oriented gradient (HOG), and spatial−temporal interesting points (STIPs). Some traditional machine learning techniques, including support vector machines (SVMs), conditional random fields (CRFs), hidden Markov models (HMMs), and random forests (RFs), are subsequently employed for the recognition of sign videos without gloss labels [2]. These methods can achieve good recognition performance when the signer's hand movement is clear and the background is simple. However, these methods often perform poorly in complex environments.

In recent years, an increasing number of researchers have focused on deep learning-based ISLR technologies. These technologies can be divided into two categories according to the modalities of the data. The first category consists of video-based methods that learn from RGB(-D) videos directly. The RGB video provides a rich representation of hand shape, movement, and other nonmanual features, such as facial expressions and body postures. These methods usually employ various CNNs to extract framewise features of sign language, such as ResNet, I3D, and S3D [8,9]. After that, they often utilize other deep learning techniques to capture temporal features, such as recurrent neural networks (RNNs), long short-term memory (LSTM) or transformers [10-13].

The second category is known as pose-based methods [14-17]. Pose-based methods are based on skeleton data, which are, in fact, the positions of the joints in the body. The skeleton data can be extracted from RGB videos via several pose estimation tools, such as OpenPose, MediaPipe and MMPose. Many studies employ transformers or graph neural networks to capture spatial–temporal features from pose information. Compared with video-based methods, pose-based methods are more robust to noise and background clutter in sign videos; moreover, they have one or two orders of magnitude fewer parameters. However, in most cases, the pose-based ISLR methods perform worse than the video-based methods do. Moreover, their performance relies greatly on pose estimation tools.

To improve the recognition accuracy, many studies have further explored fusion methods that fuse multiple input modalities. For example, Zuo et al. [18] employed S3D to extract features from RGB videos and pose data, improving the recognition accuracy. Jiang et al. [19] proposed the SAM−SLR model, which uses a multimodal approach to learn the GCN and 2D-CNN for sign language recognition and finally makes predictions through an ensemble method. In addition, some recent works have employed new technologies, such as BERT [20,21] and transfer learning [22,23], and have achieved remarkable recognition accuracy. Although the academic community has made great progress in ISLR, many problems remain unsolved, such as hand occlusion.

**2.2 Datasets for ISLR**

Since sign language is a typical low-resource language, it is not easy to construct a large-scale dataset for the study of ISLR. Nevertheless, researchers have developed more than 20 ISLR datasets. In this section, we briefly introduce several publicly available datasets that are popular in ISLR studies. MSASL [24] and WLASL [25] are two popular ISLR datasets of American Sign Language (ASL), both of which are obtained from the Internet. They contain 1000 glosses and 2000 glosses in ASL respectively. The MSASL dataset includes 25513 sign videos performed by 222 signers, and the WLASL dataset includes 21083 sign videos performed by 119 signers. However, both methods provide only a single perspective of the signers; thus, it is difficult to handle the hand occlusion issue. AUTSL is a dataset of

Turkish Sign Language (TSL) [26] that contains 226 signs performed by 43 different signers. There are 38336 sign videos in total. The videos are recorded using Microsoft Kinect v2 in RGB, depth and skeleton formats. Since the AUTSL dataset contains depth data, it is beneficial for solving the hand occlusion issue. However, depth data are too space-consuming and expensive for most application scenarios. NMFs-CSL [27] and DEVISIGN-L [7] are two ISLR datasets of Chinese Sign Language (CSL) that contain a large lexicon of 2000 glosses and 1067 glosses, respectively. The NMFs–CSL dataset contains 32010 sign videos recorded by 10 signers in a standard studio. The DEVISIGN-L dataset includes 24 thousand sign videos recorded by 8 signers using depth cameras and is recorded in a nonstandard studio environment. In addition to the above datasets, the academic community has also contributed many datasets for the study of ISLR, such as BSL-1K, LSA64, and AUSLAN. However, these datasets contain few glosses, and they all provide sign videos from a single perspective. Recently, several works have addressed the development of multiview sign language datasets, such as How2Sign [28] and MM-WLAuslan [29], but none of them cover the entire sign language vocabulary.

Unlike existing datasets, the proposed NationalCSL-DP dataset is a large-scale dual-view ISLR dataset. It contains all the sign motions in the Chinese National Sign Language (CNSL), which consists of 6707 glosses. It is composed of 134140 sign videos recorded by 10 signers using RGB cameras and is recorded in a standard studio environment. Each sign video was recorded from two vertical viewpoints, namely, the front side and the left side. Our aim is to solve the hand occlusion problem through the feature fusion of two perspectives and to address the ISLR issue based on real sign language vocabulary.

**2.3 Multiview action recognition**

Multiview action recognition enhances action recognition systems by utilizing multiple camera perspectives, addressing challenges of single-view systems such as occlusions, viewpoint variations, and background clutter [30,31]. The current dominant approaches consist of skeleton, RGB, or multimodal methods. Skeleton-based methods are the most common approach for multiview action recognition because of their quality representation of motion, their ability to remove irrelevant information such as background and clothing, and the widespread availability of accurate skeleton ground truth labels. Chen et al. [32] introduced a channelwise topology refinement graph convolution method that optimizes the capture of skeleton joint relationships for skeleton-based action recognition. Furthermore, Shi et al. [33] focused on adapting the model size to the number of joints and specific scenarios, improving the model's practicality. Subsequently, Song et al. [34] developed an end-to-end spatial–temporal attention model from skeleton data, which enhanced recognition performance by prioritizing the most relevant temporal features. RGB-based action recognition is sparser in the literature because of its lack of 3D structure. Wang et al. [35] explored generative methods for multiview action recognition, further advancing the field. Das et al. [36] studied video–pose embedding to improve action representation. Cheng et al. [37] proposed a cross-modality interactive learning method that aggregates spatial–temporal information for multiview action recognition. In addition to unimodal approaches, multimodal learning has shown promising results because of the additional quantity and modes of information provided during training. Bruce et al. [38] proposed a multiview multimodal approach for action recognition named MMNet, in which skeleton and RGB-based features are fused in a complementary manner to learn action representations better. Moreover, Li et al. [39] delved into unsupervised view-invariant learning, enhancing recognition across different viewpoints. Das and Ryoo [40] worked on self-supervised video representations for recognizing actions from unseen viewpoints. Ji et al. [41] employed attention transfer to refine model performance by focusing on critical features, and Bian et al. [42] introduced a global–local contrastive method for skeleton-based recognition, improving model robustness. Typically, they use multiview data as input and integrate multiview features for prediction. Inspired by the above multiview tasks, we study dual-view ISLR in this paper. As verified in our experiments, dual-view features containing useful information prove to be effective for boosting performance.

### III. NationalCSL-DP dataset

**3.1 Description**

In this section, we provide some descriptions of the NationalCSL-DP datasets, including the participants, the source of the vocabulary and the video-recording environment and process.

**Participant:** For the development of the NationalCSL-DP dataset, 10 participants were recruited, including 2 males and 8 females, with a mean age of 19.82±0.28 years. Among them, 8 were deaf students, and 2 were hearing students, all of whom were highly proficient in CNSL. These participants were in charge of three tasks. The first task was to eliminate certain glosses with identical sign motions from the CNLS vocabulary. For example, "步枪" (musket) and "射击" (shoot) are two glosses with the same sign motion; thus one gloss was retained while the other was excluded. This was done to ensure that each sign word in the dataset had a distinct motion to enhance the uniqueness and clarity of the data. The second task was to record sign videos of all glosses in the vocabulary list. During this process, at least one sign language expert was present onsite to supervise the video-recording work. This expert's presence guaranteed that the recordings adhered to the correct sign language norms and standards to maintain the quality of the video data. After the videos were recorded, all the participants annotated the sign videos. Specifically, they assigned a gloss to each sign video. However, given that this annotation process was prone to errors, which will be elaborated, they carried out cross-verification. This step was essential to ensure the correctness and high quality of the dataset, minimizing potential mistakes in the annotation.

**Vocabulary list:** The newly released CNSL vocabulary list officially encompasses 8214 sign words. Before video recording, three types of sign glosses need to be considered. The first type consists of sign glosses that are distinct in meaning yet share the same hand motion, e.g., "步枪" (musket) and "射击" (shoot). In such cases, according to their order in the vocabulary list, we excluded the latter and retained the former. After this process, the clipped vocabulary list contains 6707 sign words. The second type involves sign glosses that have identical glosses but different meanings and hand motions. Taking "安全

带" (safety belt) as an example, this gloss has two distinct interpretations: the safety belt used in aircraft and that used in cars. To differentiate between them, we appended a suffix "1-X" to the original gloss label. As a result, we have "安全带 1-1" and "安全带 1-2". The third type concerns sign words with the same meaning and gloss but different hand motions. An example is "发现" (discovery). We added the suffix "2-X" after the gloss label. For example, "发现" was annotated as "发现 2-1" and "发现 2-2", which correspond to the sign gloss "discovery" in the northern and southern parts of China, respectively.

**Dataset development:** The videos were recorded in a supervised environment with two green-screen studios. Each of these studios was furnished with two high-definition RGB cameras. One camera was positioned on the front side of the signer, while the other was placed on the left side. This setup was designed to capture the sign motions from dual perspectives. All the cameras were configured to record videos at a resolution of 1920×1080 pixels and a frame rate of 50 frames per second. Each gloss in the vocabulary list was signed by ten signers. During the recording process, as soon as a signer started signing, the two cameras began to record synchronously. A team consisting of two assistants and at least one sign expert was present. The assistants were responsible for camera operation, and the expert monitored the signer's motions to guarantee that the videos met the required quality standards. All the signers were required to wear black T-shirts. They were also given specific instructions regarding their body position. At the beginning and end of each video, they placed their arms naturally at their sides. Additionally, it was essential that the upper body above the knee, especially the arms, remained within the video frame throughout the entire signing process. This ensured that all relevant sign language gestures were captured. Furthermore, a rigorous cross-check was performed to ensure the correctness and quality of the dataset. The entire video recording process (including annotation and cross-check) spanned approximately 5 months.

Finally, the NationalCSL-DP dataset contains a total of 134140 sign videos, with the total duration of the videos exceeding 88 hours. To our knowledge, this is the first ISLR dataset for CSL that provides dual-view RGB videos and covers the complete glosses in the CNSL vocabulary. Figure 2 shows an overview of the proposed dataset.

### 3.2 Process and arrangement of the dataset

To enhance usability, we preprocessed the raw sign videos in the dataset. Each original sign video was transformed into a series of image frames. These frames were sampled at a consistent rate from the raw sign videos captured from two perspectives. These images were named on the basis of their sequential order, such as 00001.jpg, 00002.jpg, and 00003.jpg. Furthermore, we applied central cropping to the images. This operation ensured that the pixel dimensions in both the horizontal and vertical directions remained equal. After that, the resolutions of all the image files were adjusted to 256×256. This standardized resolution enables better compatibility and processing in the ISLR task.

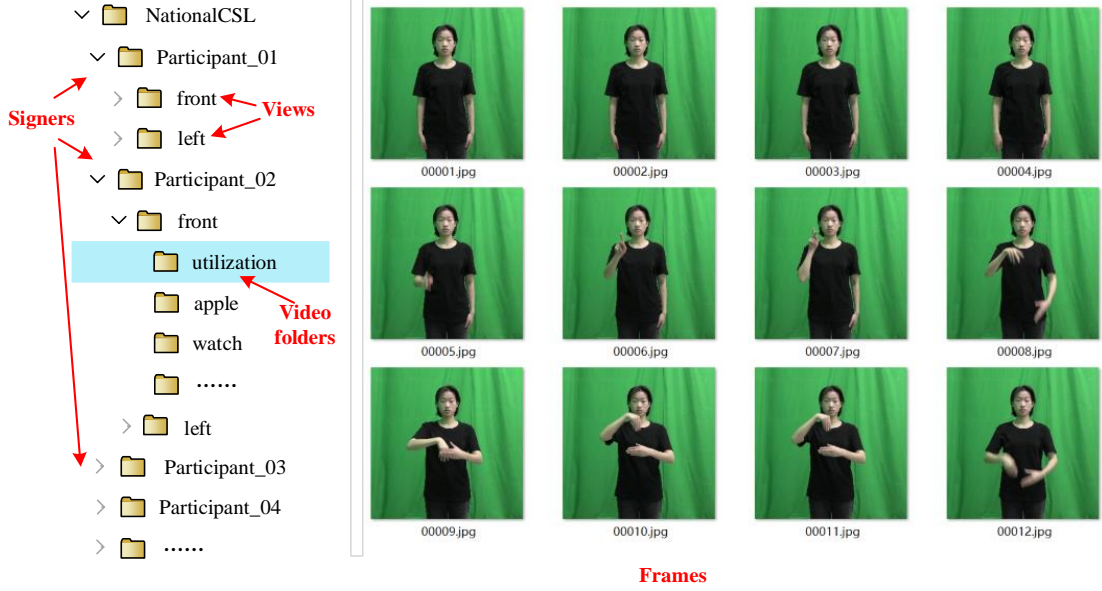

Figure2. Overview of the NationalCSL-DP dataset

Similar to the WLASL dataset, we created five subsets from the original dataset, each containing a distinct number of glosses. This approach was designed to offer deeper insights into the challenges inherent in the ISLR task and the scalability of sign recognition algorithms. Specifically, we randomly selected *K* glosses from the vocabulary list, where *K* = {200, 500, 1000, 2000, 6707}. We subsequently organized these selected glosses into five separate subsets. These subsets were named NationalCSL200, NationalCSL500, NationalCSL1000, NationalCSL2000, and NationalCSL6707. This systematic division allows for more targeted experimentation and analysis within the context of sign language research, enabling us to evaluate how different levels of data complexity impact the performance of sign recognition systems.

The NationalCSL-DP dataset, as well as a gloss-filename mapping file and the partition files, can be downloaded from http://ai.lsnu.edu.cn/info/1009/1830.htm. Each sign video is named according to the order number of its gloss word in CNSL. Furthermore, we performed central cropping on the sign videos and scaled them to a resolution of 512×512.

## IV. CNN–transformer model with different fusion strategies

In this work, a CNN−transformer model is proposed to recognize dual-view isolated sign videos. Specifically, a CNN is used to extract features as a spatial backbone $S$ is used for each frame, and a transformer encoder $T$ with an attention mechanism is used as a sequence processing network. For a front-view isolated sign video with $T$ frames, $X_{front} = \{X_t\}_{t=1}^{T} \in R^{T \times 3 \times H \times W}$, and for a left-view video, $X_{left} = \{X_t\}_{t=1}^{T} \in R^{T \times 3 \times H \times W}$ where $H$ and $W$ represents the height and width of each frame, respectively. The proposed CNN−transformer model aims to map the input dual-view videos into a gloss. Furthermore, two spatial extractors first process input frames into framewise features $p_{front} = \{p_t\}_{t=1}^{T} \in R^{T \times d_s}$ and $p_{left} = \{p_t\}_{t=1}^{T} \in R^{T \times d_s}$, where $d_s$ is the spatial representation dimension, according to the following equation:

$$p_{front} = S_{front}(X_{front}) \quad (1)$$
$$p_{left} = S_{left}(X_{front}) \quad (2)$$

For the proposed dual-view NationalCSL-DP, the fusion strategies include early fusion $F_{early}$, late fusion $F_{late}$ and a simple but effective fusion strategy named plus fusion $F_{plus}$, as shown in Figure 3. Moreover, early fusion and late fusion both utilize front and left videos during training and prediction. However, plus fusion trains the front view model separately from the left view model and performs fusion during the prediction phase.

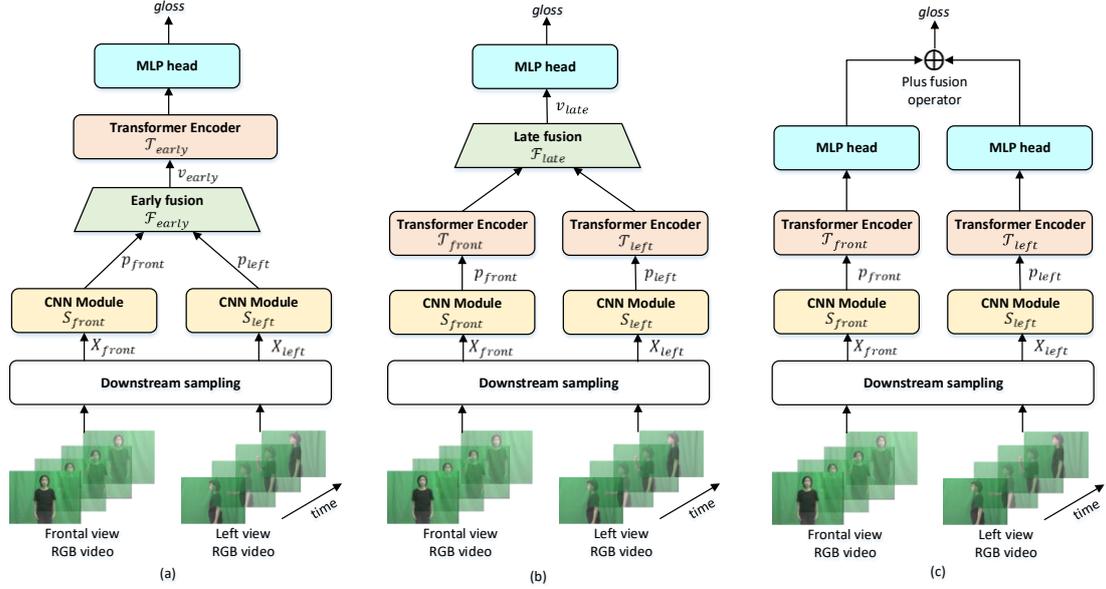

Figure 3. Proposed CNN–transformer model with different fusion strategies

For early fusion $F_{early}$, as shown in Figure 3(a), the features extracted by the spatial backbones from the front and left views are fused as $v_{early} \in R^{T \times d_{early}}$, where $d_{early}$ represents the attention dimension as follows:

$$v_{early} = F_{early}(p_{front}, p_{left}) = Linear\left(Concat(p_{front}, p_{left})\right) \quad (3)$$

where $Linear$ refers to a linear transformation and where $Concat$ is a concatenation of $p_{front}$ and $p_{left}$. For recognition, $v_{early}$ is input into the transformer encoder $T_{early}$ with an MLP head $H_{early}$ after pooling $T$ frames, as follows:

$$gloss = H_{early}\left(\frac{1}{T}\sum_{i=1}^{T} T_{early}(v_{early})\right) \quad (4)$$

In the case of late fusion $F_{late}$ illustrated in Figure 3(b), the video features extracted by the transformer encoders are fused as $v_{late} \in R^{T \times d_{late}}$, where $d_{late}$ represents the attention dimension by the following equation:

$$v_{late} = F_{late}\left(T_{front}(p_{front}), T_{left}(p_{left})\right) = Linear\left(Concat\left(T_{front}(p_{front}), T_{left}(p_{left})\right)\right) \quad (5)$$

where $Linear$ refers to a linear transformation and where $Concat$ is a concatenation function. Conclusively, the gloss is recognized by the following equation:

$$gloss = H_{late}\left(\frac{1}{T}\sum_{i=1}^{T}(v_{late})\right) \quad (6)$$

When considering plus fusion in Figure 3(c), during the training phase, we train the recognition model $M_{view}$, including a front-view recognition model $M_{front}$, independently from a left-view model $M_{left}$ as follows:

$$M_{view}(X_{view}) = H_{view}\left(T_{view}(S_{view}(X_{view}))\right) \tag{7}$$

Finally, the font view and the left view features are fused at the prediction phase as follows:

$$gloss = Softmax(M_{front}(X_{front}), M_{left}(X_{left})) \tag{8}$$

## V. Experiments

**5.1 Experimental design**

This section verifies the effectiveness of the proposed NationalCSL-DP dataset as well as the baseline. The experiments are aimed at two aspects. First, we verify whether the recognition accuracy of ISLR can be significantly improved by training on a dual-view sign video dataset. Second, three fusion strategies are compared, and a powerful baseline method is selected for subsequent research. For the first objective, we selected a skeleton-based ISLR method, named SL-GCN [19], to use in the comparative experiment. We employed the proposed CNN–transformer and SL-GCN methods combined with the plus fusion strategy and conducted experiments on single-view and dual-view sign language video datasets. For the second objective, the proposed CNN–transformer method was combined with three different fusion strategies for experimental comparison. The experiments were carried out on a server equipped with 8 NVIDIA GeForce RTX 4090 GPUs. The experimental environments were CentOS 7.0 and PyTorch 1.8.

**5.1 Evaluation metric**

The top-*k* accuracy is defined as the proportion of test samples for which the correct class is among the top-*k* classes predicted by the model. This measure is particularly suitable for tasks such as ISLR, which encompass a broad spectrum of possible outcomes. The formula for the top-*k* accuracy is given by

$$Top - k \; accuracy = \frac{1}{N}\sum_{i=1}^{N} \mathrm{I}(y_i \in \hat{Y}_i^k) \tag{9}$$

Here, $N$ represents the total number of samples (a.k.a. sign videos), and the indicator function $\mathrm{I}$ returns a value of 1 if the true label $y_i$ for the $i_{th}$ sample falls into the subset of the top-*k* predicted labels $\hat{Y}_i^k$, which are output by the inference model.

**5.2 Implementation details**

This section provides the details of the experimental setup used in this study. We extracted 16 frames from the central portion of each video, with a spacing of 5 frames between each, thereby creating an effective temporal receptive field spanning 80 frames. Across all the experiments, we maintained consistent hyperparameter configurations. For training, we employed categorical cross-entropy as the loss function and paired it with the SGD optimizer with a batch size of 32. The training process continued until the validation loss ceased to decline over a span of 5 epochs. We then selected the model checkpoint from the epoch that resulted in the lowest loss as our final model for evaluation. For 2D feature extraction, we utilized a ResNet-34 model pretrained on ImageNet, which yields a 512-dimensional feature vector for each frame. To process the sequence in the latent space, we applied 4 layers of 8-head attention mechanisms. In each experiment, the size of the embedding within the self-attention mechanism was set to 512.

**5.3 Single view vs. dual view**

In this study, we employ a SLGCN, which is grounded in a GCN via the skeleton modality, and the CNN–transformer model proposed in this work is based on computer vision via the RGB modality. As shown in Table 1, the accuracies of dual-view isolated videos are notably greater than those of single-view videos.

Table 1: Accuracy of single-view vs. dual-view methods (%)

| Datasets | Metrics | Left view | | Front view | | Dual-view | |
|---|---|---|---|---|---|---|---|
| | | SL–GCN | CNN–Transformer | SL–GCN | CNN–Transformer | SL–GCN | CNN–Transformer |
| NationalCSL200 | Top-1 | 65.00 | 68.50 | 71.50 | 76.50 | 81.50 | **85.50** |
| | Top-5 | 83.50 | 87.00 | 92.50 | 93.00 | 96.00 | **98.00** |
| | Top-10 | 89.00 | 90.50 | 96.00 | 97.00 | 97.50 | **98.00** |
| NationalCSL500 | Top-1 | 55.60 | 60.40 | 71.20 | 76.60 | 80.20 | **84.00** |
| | Top-5 | 82.80 | 85.20 | 89.40 | 92.80 | 91.20 | **94.20** |
| | Top-10 | 87.20 | 89.40 | 93.00 | 94.40 | 94.80 | **96.60** |
| NationalCSL1000 | Top-1 | 55.70 | 59.30 | 69.30 | 72.80 | 75.80 | **80.70** |
| | Top-5 | 80.10 | 83.70 | 86.90 | 90.10 | 90.50 | **93.80** |
| | Top-10 | 85.00 | 88.70 | 91.10 | 93.50 | 94.70 | **96.80** |
| NationalCSL2000 | Top-1 | 51.65 | 55.30 | 70.95 | 73.25 | 77.10 | **79.30** |
| | Top-5 | 75.95 | 79.55 | 87.00 | 88.50 | 89.45 | **92.85** |
| | Top-10 | 83.60 | 86.05 | 90.25 | 92.40 | 94.40 | **95.45** |
| NationalCSL6707 | Top-1 | 39.92 | 40.99 | 62.91 | 64.34 | 66.17 | **69.61** |
| | Top-5 | 64.87 | 68.11 | 79.31 | 83.66 | 84.93 | **88.92** |
| | Top-10 | 73.77 | 76.32 | 84.88 | 88.41 | 88.72 | **92.93** |

The table shows that regardless of whether the CNN–transformer model or the SL–GCN model is used, the recognition accuracies achieved on the dual-view dataset are markedly superior to those obtained on the single front-view dataset. Specifically, in the comparison between the dual-view datasets and front-view datasets, the SL–GCN model achieved improvements of 10.00%, 9.00%, 6.50%,

6.25%, and 3.26% in the top-*1* accuracy on the NationalCSL200, NationalCSL500, NationalCSL1000, NationalCSL2000 and NationalCSL6707 datasets, respectively, whereas the CNN−transformer model achieved improvements of 9.00%, 7.40%, 7.90%, 6.05%, and 5.27% in the top-*1* metric on the five datasets. In addition, not only does the top-*1* accuracy improve, but the top-5 and top-10 accuracies also improve on the dual-view NationalCSL-DP compared with the single-view accuracy. In conclusion, the proposed dual-view NationalCSL-DP is beneficial for increasing the recognition accuracy of ISLR. This proves the effectiveness of the proposed NationalCSL-DP dataset.

In addition, the CNN−transformer model outperforms the SL−GCN model in terms of all the metrics on the five datasets, including the left-view datasets, the front-view datasets, and the dual-view datasets. Compared with those of the SL−GCN model, the top-1 accuracies of the CNN−transformer model on the NationalCSL200, NationalCSL500, NationalCSL1000, NationalCSL2000 and NationalCSL6707 datasets are 4.00%, 3.80%, 4.90%, 2.20% and 3.44%, respectively. For NationalCSL6707, which contains at least three times as many glosses as DEVISIGN and WLASL do, the CNN−transformer model achieves a top-*1* accuracy of 69.61%. This proves that the proposed CNN−transformer model can be regarded as a strong baseline for the NationalCSL-DP dataset.

**5.4 Comparison of different fusion strategies**

When addressing the proposed dual-view NationalCSL-DP, we explored various fusion strategies to enhance the extraction of dual-view features. In this study, we employed three popular strategies—early fusion, late fusion, and plus fusion—along with our proposed baseline model, as detailed in Table 2.

Our findings indicate that fusing features after the sequence processing network hinders performance in such dual-view ISLR tasks. For example, compared with the top-*1* accuracies obtained on the single front-view datasets, the top-*1* accuracies of the late fusion strategy decrease by 6.50%, 5.60%, 8.60%, 13.60%, and 15.84% on the NationalCSL200, NationalCSL500, NationalCSL1000, NationalCSL2000, and NationalCSL6707 datasets, respectively. In the case of early fusion, integrating features following spatial extraction offers limited benefits to dual-view ISLR. Compared with the top-*1* accuracies obtained on the single front-view datasets, the top-*1* accuracies improved by a mere 1.40% on NationalCSL200 and remained stable for NationalCSL500 and NationalCSL1000, but the top-*1* accuracies decreased by 0.80% and 8.94% on NationalCSL2000 and NationalCSL6707, respectively. The plus fusion method significantly increases the accuracy: the top-*1* accuracy increases by 9.00%, 7.40%, 7.90%, 6.05%, and 5.27% on the NationalCSL200, NationalCSL500, NationalCSL1000, NationalCSL2000, and NationalCSL6707 datasets, respectively. This indicates that on the NationalCSL-DP dataset, adopting simple early fusion or late fusion strategies cannot extract complementary visual features, thereby failing to improve the recognition accuracy. However, training two recognition models separately on the basis of the two views of the NationalCSL-DP dataset and then performing fusion at the recognition stage can yield excellent results. **Since our aim is to provide a strong baseline for the proposed NationalCSL-DP dataset, we did not study why the simple early fusion and late fusion strategies are**

**ineffective. This will be left for future research.**

Table 2: Accuracy of different fusion strategies (%)

| Datasets | Metric | Fusion strategy | | |
|---|---|---|---|---|
| | | Late fusion | Early fusion | Plus fusion |
| NationalCSL200 | Top-*1* | 70.00 | 76.50 | **85.50** |
| | Top-*5* | 90.00 | 95.00 | **98.00** |
| | Top-*10* | 93.00 | 97.00 | **98.00** |
| NationalCSL500 | Top-*1* | 71.00 | 78.00 | **84.00** |
| | Top-*5* | 89.00 | 91.00 | **94.20** |
| | Top-*10* | 93.00 | 94.00 | **96.60** |
| NationalCSL1000 | Top-*1* | 64.20 | 72.80 | **80.70** |
| | Top-*5* | 86.20 | 89.40 | **93.80** |
| | Top-*10* | 90.00 | 92.90 | **96.80** |
| NationalCSL2000 | Top-*1* | 59.65 | 72.45 | **79.30** |
| | Top-*5* | 82.10 | 86.85 | **92.85** |
| | Top-*10* | 87.50 | 90.85 | **95.45** |
| NationalCSL6707 | Top-*1* | 48.50 | 55.40 | **69.61** |
| | Top-*5* | 73.55 | 77.81 | **88.92** |
| | Top-*10* | 80.66 | 83.61 | **92.93** |

## VI. Conclusions

In this work, we release a new dual-view dataset named NationalCSL-DP that includes 131K+ high-quality isolated sign videos that cover a total of 8124 sign words in the CNSV and show front and left views. Many effective steps were taken to ensure the validity of the dataset. First, all 10 participants, including males and females, deaf students and hearing students, were highly proficient in CNSL to avoid unnecessary biases and enhance the uniqueness and clarity of the data. Second, we carried out strict supervision over the dataset construction process, including video recording and annotation, to ensure the quality of the dataset. Furthermore, NationalCSL-DP was split into NationalCSL200, NationalCSL500, NationalCSL1000, NationalCSL2000, and NationalCSL6707 with different scales, allowing for more targeted experimentation and analysis within the context of sign language research.

In addition, a CNN−transformer model is proposed to evaluate the effectiveness of the NationalCSL-DP dataset. Three popular fusion strategies, i.e., early fusion, late fusion and plus fusion, are employed in the model. We conducted extensive experiments to evaluate the effectiveness of the proposed dataset and the model. The results show that (1) the proposed dual-view NationalCSL-DP dataset is beneficial for increasing the accuracy in ISLR. (2) The simple early fusion and late fusion strategies cannot extract complementary visual features from the sign videos in the NationalCSL-DP dataset, thereby failing to improve the recognition accuracy on the dataset. (3) The proposed CNN−transformer model can be regarded as a strong baseline for the NationalCSL-DP dataset.

## Contribution

Siyuan Jing was in charge of dataset development, experimental design and manuscript writing; Guangxue Wang designed and implemented the proposed baseline and participated in manuscript writing; Haoyang Zhai was in charge of implementing SL−GCN. Qin Tao and Jun Yang managed the datasets and assisted with the experiments. Bing Wang participated in the discussion. Peng Jin was the director of the study.

## Acknowledgments

This work is supported by the Humanities and Social Sciences Project of the Ministry of Education (Grant No. 23YJA740013); the Sichuan Natural Science Foundation (Grant No. 2024NSFSC0520); the General Project of the Philosophy and Social Sciences Foundation of Sichuan Province (Grant No. SCJJ24ND127); and the Research Cultivation Project of Leshan Normal University (Grant No. KYPY2024-0002).

## Data availability statement

The NationalCSL-DP dataset can be downloaded from http://ai.lsnu.edu.cn/info/1009/1830.htm. The dataset is released under the CC-BY license.

## Compliance with ethical standards

**(1) Conflicts of interest:** The authors declare that they have no conflicts of interest.

**(2) Ethical approval:** All participants provided informed consent forms for the sharing of their identity information and signed agreements to consent to participate in the construction of the NationalCSL-DP dataset, as well as to allow the dataset to be published, including but not limited to academic journals and online databases. The Ethical Review Board (ERB) of Leshan Normal University reviewed our ethical review application, as well as the informed consent forms and agreements of all participants regarding the sharing of identity information as well as the dataset publication. Finally, permission was granted by the ERB of LSNU for the open publication of the dataset, including manuscript submission and dataset release (Ethical Review Number: LNU-KYLL2025-02-15).